\documentclass{bmvc2k}

\title{Where are the Masks: Instance Segmentation with Image-level Supervision}

\addauthor{Issam H. Laradji}{issamou@cs.ubc.ca}{12}
\addauthor{David Vazquez}{dvazquez@elementai.com}{1}
\addauthor{Mark Schmidt}{schmidtm@cs.ubc.ca}{2}

\addinstitution{
 Element AI\\ Montreal, Canada
}
\addinstitution{
 University of British Columbia\\ Vancouver, Canada
}

\runninghead{Laradji, Vazquez, \& Schmidt}{Where are the Masks}

\usepackage{times,epsfig,graphicx,amsmath,amssymb}
\usepackage{times,graphicx,amsmath,amssymb,booktabs,tabulary,multirow,overpic,bbm,amssymb}

\newcolumntype{x}[1]{>{\centering\arraybackslash}p{#1pt}}

\newlength\savewidth\newcommand\shline{\noalign{\global\savewidth\arrayrulewidth
  \global\arrayrulewidth 1pt}\hline\noalign{\global\arrayrulewidth\savewidth}}
\newcommand{\tablestyle}[2]{\setlength{\tabcolsep}{#1}\renewcommand{\arraystretch}{#2}\centering\footnotesize}
\makeatletter\renewcommand\paragraph{\@startsection{paragraph}{4}{\z@}
  {.5em \@plus1ex \@minus.2ex}{-.5em}{\normalfont\normalsize\bfseries}}\makeatother

\usepackage{algorithm} 
\usepackage{algpseudocode}  

\begin{document}

\maketitle

\begin{abstract}
A major obstacle in instance segmentation is that existing methods often need many per-pixel labels in order to be effective. These labels require large human effort and for certain applications, such labels are not readily available. To address this limitation, we propose a novel framework that can effectively train with image-level labels, which are significantly cheaper to acquire. For instance, one can do an internet search for the term "car" and obtain many images where a car is present with minimal effort. Our framework consists of two stages: (1) train a classifier to generate pseudo masks for the objects of interest; (2) train a fully supervised Mask R-CNN on these pseudo masks. Our two main contribution are proposing a pipeline that is simple to implement and is amenable to different segmentation methods; and achieves new state-of-the-art results for this problem setup. Our results are based on evaluating our method on PASCAL VOC 2012, a standard dataset for weakly supervised methods, where we demonstrate major performance gains compared to existing methods with respect to mean average precision.

\end{abstract}

\section{Introduction}
\label{sec:introduction}
The recent progress in Deep Neural Networks (DNNs) and segmentation frameworks has given us major improvements in the task of instance segmentation~\cite{he2017mask, chen2017masklab}. Their success was demonstrated in various applications such as autonomous driving~\cite{cordts2016cityscapes}, scene understanding~\cite{lin2014microsoft, everingham2010pascal}, and medical imaging~\cite{pohle2001segmentation, Konopczynski2018InstanceSO}. Nonetheless, these methods require  a large number of training data with per-pixel labels, or labels which distinguish between object categories and instances in the image. As acquiring them is often prohibitively expensive, the effectiveness of these methods is limited to a small range of datasets and object categories.

Many weakly supervised methods emerged to overcome the need for per-pixel labels. Instead, they only require weaker labels ranging from bounding boxes~\cite{khoreva2017simple}, scribbles~\cite{lin2016scribblesup}, and image-level~\cite{Zhou2018PRM, cholakkal2019object} annotations. This makes acquiring datasets a significantly more scalable endeavour. According to \citet{bearman2016s}, it requires 20 sec/img to acquire image-level labels (which are labels that only indicate whether an object class appears in an image) for PASCAL VOC~\cite{everingham2010pascal}, compared to 239.7 sec/img for acquiring per-pixel labels. To date, only two weakly supervised methods address instance segmentation with image-level labels, making our work one of the few that tackles a relatively unexplored research area.

Perhaps the first work to address this problem setup is PRM~\cite{Zhou2018PRM}. It trains a classifier which can then identify local regions belonging to different objects of the same category. It extends CAM-based methods~\cite{selvaraju2017grad, ahn2018learning} by not only identifying large regions where objects are vaguely located, but also identifying peaks that represent the specific locations of the object instances. At test time, the trained PRM obtains the object masks in two steps. First, it uses the gradient with respect to the input to get a rough mask of the objects using a process called peak backpropagation. This results in a peak response map. Then, the masks in this map are replaced by the best matching proposal masks, which are generated from a pretrained object proposal method~\cite{arbelaez2014mcg, pinheiro2016sharpmask}. However, their results are much worse than that of fully supervised methods, leaving a large room for improvement (Table~\ref{tab:pascal}).

Our Weakly-supervised Instance SEgmentation method (WISE) builds on PRM by using its output pseudo masks to train a fully-supervised method, namely, Mask R-CNN~\cite{massa2018mrcnn}. Our intuition as to why this procedure is effective is that Mask R-CNN is potentially robust to noisy pseudo masks, and the noisy labels within these masks might be ignored during training as they are potentially uncorrelated. The success of such a de-noising strategy has been demonstrated in semantic segmentation and object localization~\cite{khoreva2017simple}.

We show that simple techniques for obtaining the pseudo masks lead to a surprisingly effective supervision for Mask R-CNN. We summarize our contributions as follows. (1) We present a novel framework that can effectively train a fully supervised method on pseudo mask labels obtained from image-level class labels; (2) we show that our framework is amenable to different localization and segmentation methods (for example, a density-based PRM~\cite{cholakkal2019object} can be used for localization and RetinaMask~\cite{fu2019retinamask} can be used for instance segmentation), and (3) we achieve new state-of-the-art results on a standard weakly-supervised instance segmentation benchmark.

\begin{figure}[t]
    \centering
    \includegraphics[width=0.8\linewidth]{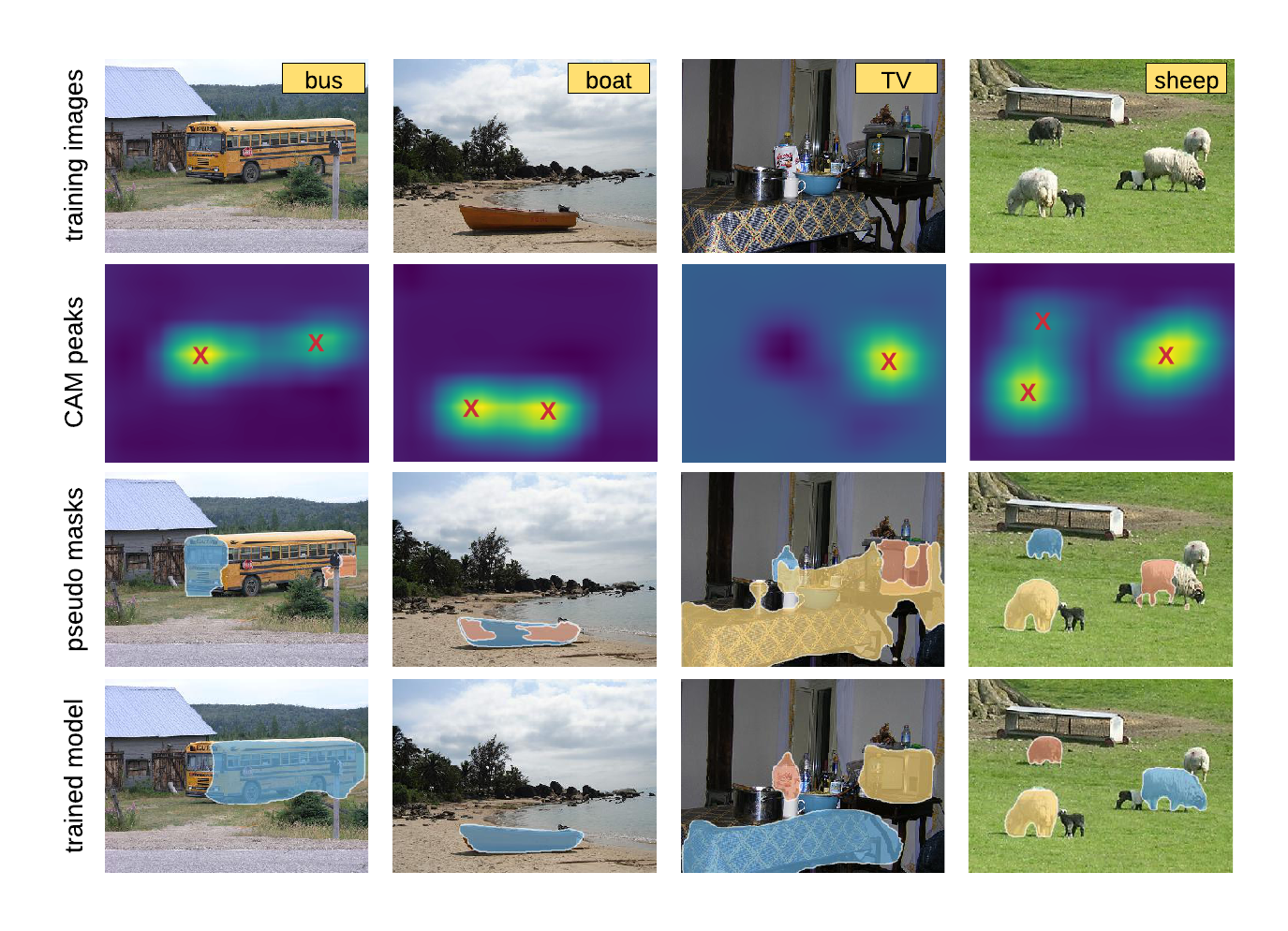}
    \vspace{-6mm}
    \caption{{\bf Framework overview.} Our Weakly-supervised Instance SEgmentation (WISE) method learns to perform instance segmentation with image-level supervision. First, a classifier is trained with a peak stimulation layer to identify peaks at which the objects are located (row 2). A proposal gallery (such as MCG~\cite{arbelaez2014mcg}) is used to obtain rough masks for the located objects, which are then used as pseudo masks to train Mask R-CNN~\cite{he2017mask} (row 3). Row 4 shows the output of a Mask R-CNN trained on the noisy pseudo mask labels.}
    \label{fig:overview}
\end{figure}

\section{Related Work}
\label{sec:relatedwork}
Instance segmentation is widely studied within the computer vision community~\cite{he2017mask, chen2017masklab, fu2019retinamask}. However, an ongoing challenge is that it is time-consuming and expensive to obtain the required per-pixel labels needed by most instance segmentation methods~\cite{everingham2010pascal, cordts2016cityscapes}. Current trends to overcome this issue leverage weaker labels (such as image-level labels) and pseudo labels obtained with the help of object proposal methods. While most of these methods are for object detection and semantic segmentation, we review them below as they are relevant.


\paragraph{Instance segmentation.}
Instance segmentation is one the most challenging tasks in computer vision. The task is to classify every object pixel into its corresponding category and distinguish between object instances~\cite{ren2017end, romera2016recurrent}. Most recent methods rely on deep networks and follow a two step procedure~\cite{he2017mask, chen2017masklab, fu2019retinamask}, where they first detect objects and then segment them. For instance, Mask-RCNN~\cite{he2017mask} uses Faster-RCNN~\cite{ren2015faster} for detection and an FCN network~\cite{long2015fully} for segmentation. In this work, we use Mask R-CNN as our fully supervised method and train it on pseudo masks instead of the costly per-pixel labels.

\paragraph{Learning with weak supervision.}
Due to the taxing task of acquiring per-pixel labels, many weakly supervised methods emerged that can leverage labels that are much cheaper to acquire~\cite{everingham2010pascal, cordts2016cityscapes}.  These labels range from bounding boxes~\cite{khoreva2017simple}, scribbles~\cite{lin2016scribblesup}, points~\cite{bearman2016s, laradji2018blobs, Laradji2019InstanceSW}, and image-level annotation~\cite{Zhou2018PRM}. Our setup considers one of the weakest forms of annotation, image-level labels.

\paragraph{Image-level labels as weak supervision.}
Acquiring image-level labels is an attractive form of annotation due to its extremely cheap cost. The annotator only needs to indicate whether a certain object class appears in an image, regardless of how many of them appear. While this form of annotation has gained steam within the research community, most of the proposed methods are for semantic segmentation~\cite{he2016deep,Tang2018WSOD,ahn2018learning}. Perhaps the lack of such research for instance segmentation is partially accounted for by the fact that instance segmentation is a more challenging task. Only recently did two works emerge for this problem setup~\cite{Zhou2018PRM, cholakkal2019object}. They extend the Class Activation Map (CAM)~\cite{zhou2016cam}, by not only identifying a heatmap that vaguely represents the regions where objects are located, but also identifying peaks of that heatmap that represent the locations of different objects. At test time, they adopt a post-processing step that matches each located object with a proposal, generated from an object proposal method. These proposals are considered as the final instance segmentation output. In contrast, we use these outputs as pseudo masks to train a fully supervised method.

\paragraph{Learning with pseudo labels.}
Our method adopts the following pipeline, generate pseudo-labels and then training a model on these labels in a fully-supervised manner. While this is novel for instance segmentation, similar approaches were used for object detection \cite{tang2017multiple} and semantic segmentation~\cite{Dai2015boxsup, Qi2016imglev, khoreva2017simple} in weakly supervised settings. However, these methods cannot be directly applied to instance segmentation, as they do not distinguish between object instances. Many such methods also rely on object proposals~\cite{hosang2016makes} to ease the task of detection~\cite{Tang2018WSOD, Bilen2016CVPR}, and segmentation~\cite{pinheiro2015image, bearman2016s, Zhou2018PRM, kolesnikov2016seed}. Object proposals are class-agnostic methods that can output thousands of object candidates per image and have progressed significantly over the last decade~\cite{uijlings2013selective, zitnick2014edge, arbelaez2014mcg, maninis2016convolutional, pinheiro2015learning, pinheiro2016sharpmask}. Similar to PRM~\cite{Zhou2018PRM} and PRM+Density~\cite{cholakkal2019object}, we also leverage object proposals to generate the pseudo masks.

\begin{figure}[t]
    \centering
    \includegraphics[width=\linewidth]{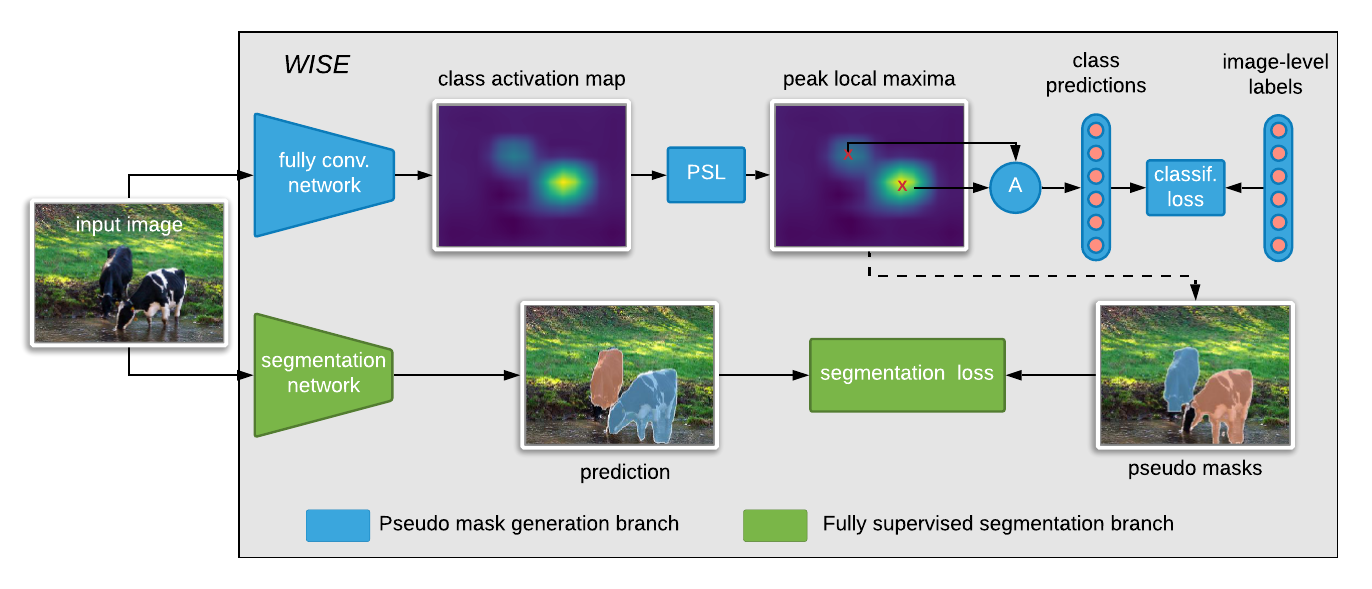}
    \vspace{-9mm}
    \caption{{\bf WISE training.} The first component (shown in blue) learns to classify the images in the dataset. The classifier first outputs a class activation map (CAM); then, obtains CAM's local maximas using a peak stimulation layer (PSL). To train the classifier, the classification loss is computed using the average of these local maximas. As the CAM peaks represent located objects, we select a proposal for each of these objects to obtain pseudo masks. The second component (shown in green) trains a Mask R-CNN on these pseudo masks.}
    \label{fig:model}
\end{figure}

\section{Proposed Method}
\label{sec:proposed_method}
Our approach to instance segmentation with image-level supervision consists of two main steps: (1) obtain pseudo masks for the training images given their ground-truth image-level labels; and (2) train a fully supervised instance segmentation method on these pseudo masks (Figure~\ref{fig:model}). In particular, this framework is based on two components: a network that obtains the pseudo masks by training a PRM~\cite{Zhou2018PRM} on the image-level labels and leveraging object proposal methods~\cite{arbelaez2014mcg}; and a Mask R-CNN~\cite{he2017mask} as a fully supervised instance segmentation method. We show the training steps of our framework in Algorithm~\ref{alg:TheAlgorithm}.

At test time, we can predict the object instance masks using the trained Mask R-CNN only, discarding the PRM component. In this setup, we are interested in segmenting $C$ classes of objects. For a training image, the image-level label is given as $Y = [y_1, y_2, ..., y_C]$, where $y_i = 1$ or $0$, indicating whether the image has an object of class $i$. We describe our components in more detail below, and also investigate a post-processing steps that can improve Mask R-CNN's final output.

\begin{algorithm}[t]
  \caption{WISE training}
  \begin{algorithmic}[1]
    \State Train a CAM-based classifier $C$ until convergence as in PRM~\cite{Zhou2018PRM};
    \While {$iter < max\_iter$}
      \State Randomly sample a training image $I$;
      \State Generate a set of proposals $P$ for $I$;
      \State Use PSL on $C$ to obtain the set of peaks $L$ for $I$;
      \State Initialize an empty list of Targets $T$;
      \For {$(i_k, j_k) \in L$}
        \State Select a proposal $(G_k, b_k)$  randomly using Eq.~\ref{eq:proposals}, it has to intersect with $(i_k, j_k) $;
        \State Add $G_k$ to list $T$;
      \EndFor
     \State Compute $\mathcal{L}(I, T, \theta)$ as in Eq.~\ref{eq:mrcnn};
     \State Update the weights for $\theta$ using back-propagation;
    \EndWhile
  \end{algorithmic}
\label{alg:TheAlgorithm}
\end{algorithm}

\subsection{Pseudo Mask Generation Branch}
We rely on PRM~\cite{Zhou2018PRM} to generate segmentation seeds that identify salient parts of the objects. These seeds help in generating pseudo masks as a source of supervision for Mask R-CNN. Following PRM's methodology, we train a CAM-based classifier which has a fully convolutional network (FCN) followed by a peak stimulation layer (PSL). As shown in Figure~\ref{fig:model}, the FCN outputs a class activation map (CAM) which specifies the class confidence at each image location. Then, PSL outputs $N^c$ local maximas of CAM within a window size $r$, namely, $L^c=\{(i_1, j_1), (i_2, j_2),...,(i_{N^c}, j_{N^c})\}$ which represents locations in the CAM for the $c$-th object class (more details in \citet{Zhou2018PRM}). In order to boost the activations of these local maximas, their average activation is first computed as, $s^c = \frac{1}{N^c}\sum_{(i_k, j_k) \in L^c}M^c_{i_k, j_k}$, where $M^c$ is the activation map corresponding to class $c$. This average is then used for binary classification, specifically the multi-label soft-margin loss~\cite{lapin2018analysis}. This classification component is trained until convergence.

We then generate the pseudo masks for the training images by using the trained classifier and an off-the-shelf object proposal method (specified as the dotted line in Figure~\ref{fig:model}). The peaks obtained from PSL, which represent object locations in the image, are replaced with proposal masks based on their ``objectness'', which are scores given by the proposal method as confidence measure for being objects. We adopt a de-noising strategy where we select a proposal randomly based on its objectness score: proposals with higher objectness are more likely to be selected. More formally, to obtain the mask for an object located at peak $(i,j)$, we first generate a set of $n$ proposals whose masks intersect with $(i,j)$, namely, $P=\{(G_1, b_1), (G_2, b_2),...,(G_n, b_n)\}$ with mask $G_k$ and objectness score $b_k$. Then, the probability of selecting a proposal mask $G_k$ is,
\begin{equation}
    P(G_k) = \frac{b_k}{\sum^n_{j=1}{b_j}}
    \label{eq:proposals}
\end{equation}

The rationale behind selecting proposals randomly is that they have common pixels that correspond to the salient parts of the located object, despite the fact that they have different objectness. While proposal masks are not originally associated with a class label, we get the object class label information from CAM and assign it to the corresponding proposals. These proposals can be used as pseudo masks to train a fully supervised instance segmentation method.


\subsection{Fully Supervised Segmentation Branch}
We can construct the segmentation labels for all the training images by using the trained pseudo mask generation branch. These are used as supervision to train a Mask R-CNN (shown as green components in Figure~\ref{fig:model}). Depending on the application, other choices of fully supervised methods can be used instead of Mask R-CNN: if the goal is to perform instance segmentation at real-time, one can consider training a YOLACT~\cite{bolya2019yolact}, and for semantic segmentation, one can consider training a DeepLab~\cite{chen2018deeplab} segmentation network.

Mask R-CNN~\cite{he2017mask} combines Faster R-CNN~\cite{ren2015faster} and FCN-based~\cite{long2015fully} methods to first detect the objects and then segment them. For an image $I$, with target pseudo masks $T$, Mask R-CNN with parameters $\theta$ is trained by optimizing the following objective function,
\begin{equation}
    \mathcal{L}(I, T, \theta) = \mathcal{L}_{\text{cls}} + \mathcal{L}_{\text{box}} + \mathcal{L}_{\text{mask}},
    \label{eq:mrcnn}
\end{equation}
where $\mathcal{L}_{\text{cls}}$ is the classification loss for the detected objects, $\mathcal{L}_{\text{box}}$ is the localization loss for the detected objects, $\mathcal{L}_{\text{mask}}$ is their segmentation loss. These terms are explained in more detail in the original Mask R-CNN paper~\cite{he2017mask}.

\begin{figure*}[t]
    \centering
    \includegraphics[width=\linewidth]{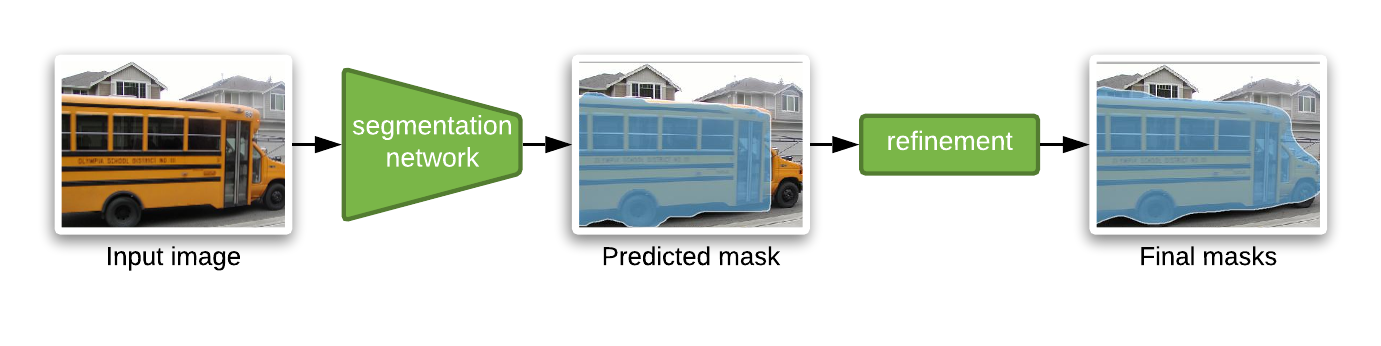}\vspace{-13mm}
    \caption{{\bf Inference.} At test time, only the trained Mask-RCNN is required to output the prediction masks in the image. As an optional refinement step, the predicted masks can be replaced with the object proposals of highest Jaccard similarity.}
    \label{fig:inference}
\end{figure*}

At test time, we can simply use the trained Mask R-CNN to predict the object masks for an unseen image. To refine these masks, we leverage the same object proposal method as that used in training. In turn, we replace each predicted object mask with the proposal of highest Jaccard similarity. Figure~\ref{fig:inference} illustrates how this refinement process can lead to a better object mask.

\section{Experiments}
\label{sec:experimental_setup}

In this section, we demonstrate the efficacy of our method by comparing it against previous methods and analyzing the pseudo masks.

\subsection{Experimental Setup}
We follow the setup by \cite{Zhou2018PRM, cholakkal2019object} for a fair benchmark, where the model only has access to an off-the-shelf proposal method and image-level labels for the training set. Also from their setup, we adopt the evaluation metric, mean average precision for Intersection-over-Union (IoU) of 0.25, 0.5, and 0.75. 

Like other works in the literature of weakly supervised methods, we perform all comparisons on the PASCAL VOC 2012 dataset~\cite{everingham2010pascal}. The dataset represents a diverse set of everyday scenes. It is divided into 1442 images for training, and 1449 images for validation. Annotators for this dataset acquired per-pixel labels for 20 objects, ranging from inanimate objects such as airplanes and bikes, and living objects such as humans and horses. However, we only use the image-level labels to train our methods.


\begin{table}[t]
    \centering
    \resizebox{0.8\columnwidth}{!}{%
    \begin{tabular}{l|x{60}|x{20}|x{20}|x{20}|x{20}}
        \multicolumn{1}{c|}{\textbf{Method}}  &
        \multicolumn{1}{c|}{\textbf{Supervision}}  &
        \multicolumn{1}{c|}{\textbf{mAP25}}  &
        \multicolumn{1}{c|}{\textbf{{mAP50}}} &
        \multicolumn{1}{c|}{\textbf{{mAP75}}} &
        \multicolumn{1}{c}{\textbf{{ABO}}}  \\\shline
        Mask R-CNN~\cite{he2017mask}           & pixel-level  & 58.9 & 51.4 & 32.4 & - \\
        DeepMask~\cite{khoreva2017simple}      & pixel-level  & -    & 41.7 & 09.7 & - \\
        \hline
        PRM~\cite{Zhou2018PRM}                 & image-level  & 44.3 & 26.8 & 09.0 & 37.6\\
        PRM+Density~\cite{cholakkal2019object} & image-level++& 48.5 & 30.2 & 14.4 & 44.3\\
        DeepMask~\cite{khoreva2017simple}      & bounding box & 39.4 & 08.1 & - & - \\
        \hline
        WISE (Ours)                            & image-level  & 48.5 & 40.4 & 22.2 & 51.3\\
        WISE+Refine (Ours)                     & image-level  & \textbf{49.2} & \textbf{41.7} & \textbf{23.7} & \textbf{55.2}\\
    \end{tabular}
    }
    \vspace{3mm}
    \caption{\textbf{PASCAL VOC 2012.} Comparison of our framework (WISE) against other methods on various levels of supervision. WISE+Refine uses the refinement step shown in Figure~\ref{fig:inference}. Mask R-CNN and DeepMask use full supervision, whereas PRM uses image-level labels. Same as WISE, PRM and PRM+Density leverage a pretrained proposal method. Requiring stronger supervision than WISE, DeepMask and PRM+Density have access to bounding box and image-level counts, respectively.}
    \label{tab:pascal}
\end{table}

\subsection{Implementation Details}
We discuss our method's procedure and parameters below. We also plan to make the code publicly available.

{\bf Network architecture.} As a common practice, we use the ResNet-50~\cite{he2016deep} that is pretrained on ImageNet~\cite{deng2009imagenet} as the backbone for PRM and Mask R-CNN. Unlike PRM, Mask R-CNN's backbone is equipped with a feature pyramid network~\cite{Lin2016FPN} that extracts features at different resolutions. The pretrained weights, along with the rest of the parameters, are then finetuned on the PASCAL VOC 2012 training set. The remaining parameters of PRM and Mask R-CNN are in the implementation details discussed in \citet{Zhou2018PRM}, and \citet{he2017mask}, respectively.

{\bf Optimization parameters.} Following the official code of Mask R-CNN, we scale its input images so that the short axis has a minimum of 800px and the long axis a maximum of 1333px. Using a single GPU of TitanX, we set our batch size as 1 and train using the SGD optimizer with a learning rate of 0.00125 for 50K iterations. This learning rate was adjusted from~\citet{he2017mask}, where they used a bigger batch size. We also augment the dataset with horizontal flips and color jittering as recommended by~\citet{deng2009imagenet}. PRM was trained as described in \citet{Zhou2018PRM}. 

\begin{table*}[t]
    \centering
    \tablestyle{4pt}{1.05}
    \resizebox{\textwidth}{!}{%
    \begin{tabular}{l|x{12}x{12}x{12}x{12}x{12}x{12}x{12}x{12}x{12}x{12}x{12}x{12}x{12}x{12}x{12}x{12}x{12}x{12}x{12}x{12}|x{12}x{12}}
        \multicolumn{1}{l|}{\textbf{Method}}  &
        \multicolumn{1}{c|}{\textbf{\rotatebox[origin=c]{90}{aero}}} &
        \multicolumn{1}{c|}{\textbf{\rotatebox[origin=c]{90}{bike}}} &
        \multicolumn{1}{c|}{\textbf{\rotatebox[origin=c]{90}{bird}}} &
        \multicolumn{1}{c|}{\textbf{\rotatebox[origin=c]{90}{boat}}} &
        \multicolumn{1}{c|}{\textbf{\rotatebox[origin=c]{90}{bottle}}} &
        \multicolumn{1}{c|}{\textbf{\rotatebox[origin=c]{90}{bus}}} &
        \multicolumn{1}{c|}{\textbf{\rotatebox[origin=c]{90}{car}}} &
        \multicolumn{1}{c|}{\textbf{\rotatebox[origin=c]{90}{cat}}} &
        \multicolumn{1}{c|}{\textbf{\rotatebox[origin=c]{90}{chair}}} &
        \multicolumn{1}{c|}{\textbf{\rotatebox[origin=c]{90}{cow}}} &
        \multicolumn{1}{c|}{\textbf{\rotatebox[origin=c]{90}{table}}} &
        \multicolumn{1}{c|}{\textbf{\rotatebox[origin=c]{90}{dog}}} &
        \multicolumn{1}{c|}{\textbf{\rotatebox[origin=c]{90}{horse}}} &
        \multicolumn{1}{c|}{\textbf{\rotatebox[origin=c]{90}{motor}}} &
        \multicolumn{1}{c|}{\textbf{\rotatebox[origin=c]{90}{person}}} &
        \multicolumn{1}{c|}{\textbf{\rotatebox[origin=c]{90}{plant}}} &
        \multicolumn{1}{c|}{\textbf{\rotatebox[origin=c]{90}{sheep}}} &
        \multicolumn{1}{c|}{\textbf{\rotatebox[origin=c]{90}{sofa}}} &
        \multicolumn{1}{c|}{\textbf{\rotatebox[origin=c]{90}{train}}} &
        \textbf{\rotatebox[origin=c]{90}{tv}} &
        \textbf{\rotatebox[origin=c]{90}{Avg.}}  \\\shline
        \scriptsize Mask R-CNN       & 71.2 &0.3 &72.2 &53.2 &29.8 &68.7 &47.3 &77.1 &13.3 &54.7 &41.0 &65.5 &51.5 &69.6 &57.8 &31.0 &46.9 &45.6 &69.7 &61.4 &51.4\\
        \hline
        \scriptsize WISE                            & 59.2 &0.6 &62.6 &38.6 &18.8 &57.3 &31.7 &66.9 &8.3 &40.5 &11.0 &55.5 &48.7 &60.2 &34.4 &24.4 &38.3 &33.1 &61.7 &56.9 &40.4\\
        \scriptsize WISE+Refine                             & 63.2 &0.3 &60.7 &39.1 &21.0 &59.4 &31.9 &68.6 &9.2 &43.1 &15.6 &58.0 &48.6 &62.3 &36.4 &21.9 &38.8 &34.3 &65.5 &56.9 &41.7
    \end{tabular}
    }
    \caption{\textbf{PASCAL VOC 2012.} Per-class comparison against the mAP$_{50}$ metric on PASCAL VOC 2012 validation set. Mask R-CNN was trained with the ground-truth per-pixel labels.}
    \label{tab:pascal_per_class}
\end{table*}

\subsection{Comparison to Previous Work} 
\begin{figure}[t]
    \centering
    \includegraphics[width=0.8\textwidth]{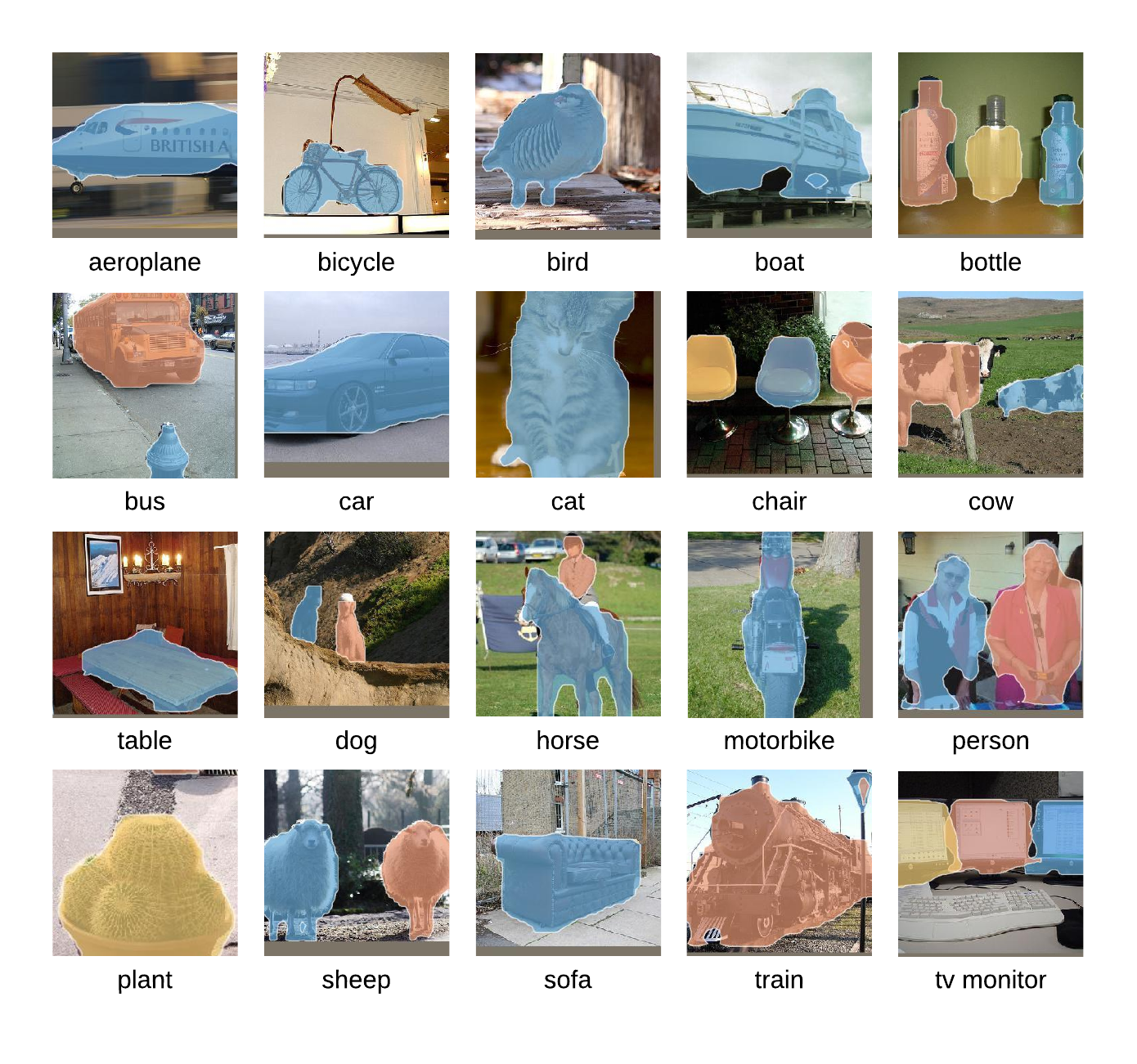}\vspace{-7mm}
    \caption{\textbf{Qualitative results.} Qualitative results of WISE on PASCAL VOC 2012 val. set. The images illustrate the predicted masks of the trained Mask R-CNN for different classes. }
    \vspace{-4mm}
    \label{fig:qualitative_results}
\end{figure}

We first quantitatively compare our approach against previous methods that use the same supervision as ours; that is, image-level labels, an object proposal method, and a ResNet-50 backbone pretrained on ImageNet. Table~\ref{tab:pascal} summarizes the results on the PASCAL VOC 2012 dataset. Our method significantly outperforms the current state-of-the-art by a large margin with respect to Average Best Overlap (ABO)~\cite{pont2015boosting}, mAP25, mAP50, and mAP75. Further, WISE without refinement also beats current state-of-the-art. Even more so, our method outperforms~\citet{cholakkal2019object} which uses slightly stronger labels than image-level. Their labels distinguish between images with 0, 1, 2-4, and 4-or more objects. Figure~\ref{fig:qualitative_results} visualizes qualitative results of WISE for each category. We further report the per-class results in Table~\ref{tab:pascal_per_class} and compare it against Mask R-CNN trained on the true masks with respect to mAP50. This illustrates that our results are competitive against fully-supervised methods.

Our method can also compete with those that use stronger supervision. Against DeepMask~\cite{khoreva2017simple}, our method outperforms two of their methods, one that uses bounding boxes as labels, and the other that uses full supervision as labels (see Table~\ref{tab:pascal}). Compared to Mask R-CNN trained on the pixel-level labels, our method still has a large room for improvement, which can be bridged by either improving the object localization component or the object proposal method.

While the overall results suggests that Mask R-CNN can effectively train from noisy, incomplete labels. The labels are noisy because the proposal masks are not perfect, and incomplete because PRM does not locate all the objects in the image. Indeed, we hypothesize that using a better object localizer such as that of \citet{cholakkal2019object} would lead to better results. But we leave that for future work. 

\subsubsection{Analysis of Pseudo masks}
We measure the generated pseudo masks performance by computing the mAP50 between the ground-truth and the generated masks. We also compute the mean absolute error to determine the number of identified objects in the images. These results are summarized in Table~\ref{tab:train_per_class}, which show that a large room for improvement is required for both metrics. Examples of the synthesized masks are
shown in Figure~\ref{fig:overview}, where one can see that the pseudo masks are not of high quality, yet the trained Mask R-CNN is able to output good masks in Figure~\ref{fig:qualitative_results}.

\begin{table*}[t]
    \centering
    \tablestyle{4pt}{1.05}
    \resizebox{\textwidth}{!}{%
    \begin{tabular}{l|x{12}x{12}x{12}x{12}x{12}x{12}x{12}x{12}x{12}x{12}x{12}x{12}x{12}x{12}x{12}x{12}x{12}x{12}x{12}x{12}|x{12}x{12}}
        \multicolumn{1}{l|}{\textbf{Metric}}  &
        \multicolumn{1}{l|}{\textbf{\rotatebox[origin=c]{90}{aero}}} &
        \multicolumn{1}{l|}{\textbf{\rotatebox[origin=c]{90}{bike}}} &
        \multicolumn{1}{l|}{\textbf{\rotatebox[origin=c]{90}{bird}}} &
        \multicolumn{1}{l|}{\textbf{\rotatebox[origin=c]{90}{boat}}} &
        \multicolumn{1}{l|}{\textbf{\rotatebox[origin=c]{90}{bottle}}} &
        \multicolumn{1}{l|}{\textbf{\rotatebox[origin=c]{90}{bus}}} &
        \multicolumn{1}{l|}{\textbf{\rotatebox[origin=c]{90}{car}}} &
        \multicolumn{1}{l|}{\textbf{\rotatebox[origin=c]{90}{cat}}} &
        \multicolumn{1}{l|}{\textbf{\rotatebox[origin=c]{90}{chair}}} &
        \multicolumn{1}{l|}{\textbf{\rotatebox[origin=c]{90}{cow}}} &
        \multicolumn{1}{l|}{\textbf{\rotatebox[origin=c]{90}{table}}} &
        \multicolumn{1}{l|}{\textbf{\rotatebox[origin=c]{90}{dog}}} &
        \multicolumn{1}{l|}{\textbf{\rotatebox[origin=c]{90}{horse}}} &
        \multicolumn{1}{l|}{\textbf{\rotatebox[origin=c]{90}{motor}}} &
        \multicolumn{1}{l|}{\textbf{\rotatebox[origin=c]{90}{person}}} &
        \multicolumn{1}{l|}{\textbf{\rotatebox[origin=c]{90}{plant}}} &
        \multicolumn{1}{l|}{\textbf{\rotatebox[origin=c]{90}{sheep}}} &
        \multicolumn{1}{l|}{\textbf{\rotatebox[origin=c]{90}{sofa}}} &
        \multicolumn{1}{l|}{\textbf{\rotatebox[origin=c]{90}{train}}} &
        \textbf{\rotatebox[origin=c]{90}{tv}} &
        \textbf{\rotatebox[origin=c]{90}{Avg.}}  \\\shline
        \scriptsize Mask R-CNN + GT     & 92.4 &15.1 &97.4 &87.9 &91.4 &94.4 &93.8 &100 &68.2 &93.4 &88.8 &97.4 &96.4 &95.3 &92.8 &89.3 &92.3 &97.7 &100 &100 &89.2\\\hline
        \scriptsize Pseudo Masks       & 24.5 &1.0 &29.1 &18.7 &11.3 &38.6 &26.6 &43.1 &8.0 &35.6 &6.1 &38.8 &46.2 &23.8 &10.7 &7.4 &35.9 &29.4 &41.6 &39.1 &25.8\\
                \scriptsize WISE      & 43.8 & 3.2 & 43.8 & 35.9 & 16.8 & 51.9 & 36.3 & 56.8 & 7.3 & 45.8 & 15.1 & 53.5 & 59.8 & 45.5 & 18.2 & 10.9 & 47.3 & 38.9 & 61.5 & 58.5 & 37.5\\
    \end{tabular}
    }
    \caption{\textbf{PASCAL VOC 2012 training set.} Comparison of the generated pseudo masks, and WISE's predicted masks with respect to mAP50. WISE was trained on a set of pseudo masks, and was able to output better masks for the same training images. Mask R-CNN + GT  was trained on the ground-truth per-pixel labels.}
    \label{tab:train_per_class}
\end{table*}

\begin{figure}[t]
    \centering
    \includegraphics[width=0.85\textwidth]{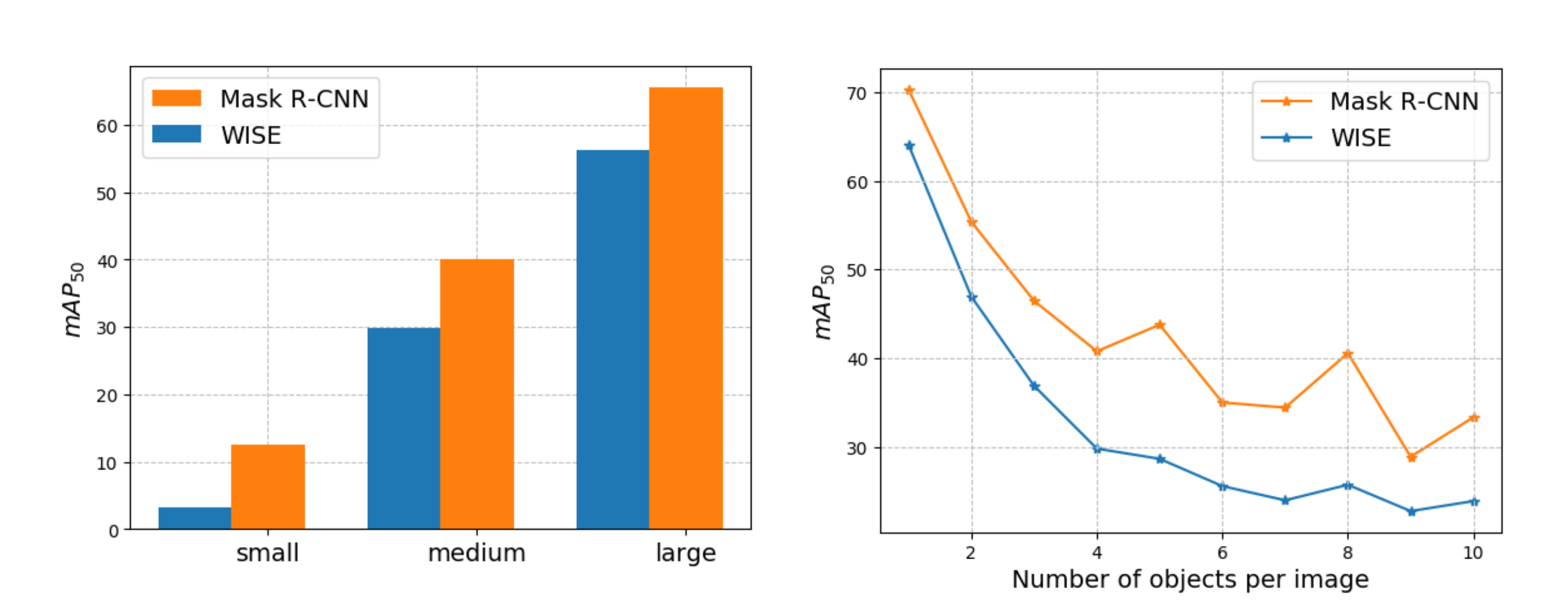}
    \vspace{-3mm}
    \caption{\textbf{Statistical Analysis.} The left figure illustrates the performance of WISE and a Mask R-CNN trained on per-pixel labels across various object sizes; and the right figure illustrates the same benchmark but across images with different number of objects.}
    \label{fig:plots}
\end{figure}

\subsubsection{Ablation Studies}
The object sizes and the number of objects in an image can have severe impact on the performance of an instance segmentation model. Figure~\ref{fig:plots} shows that WISE struggles with segmenting small objects, and when the number of objects is larger than 4. In fact, there is a heavy decline in performance when the number of objects is more than 1. More robust than WISE, a Mask R-CNN trained on per-pixel labels is able to maintain higher performance with small objects and with images with larger number of objects. In addition, such Mask R-CNN performs significantly better than WISE for small objects. This suggests that the pseudo masks trained by WISE are likely far from accurate.






\section{Conclusion}
We proposed a weakly supervised instance segmentation method that follows a two-stage pipeline for training on image-level labels. In the first stage, it uses class activation maps with a peak stimulation layer to locate the objects in the training images, and then object proposals to generate pseudo masks for these objects. In the second stage, we use Mask R-CNN to train on the pseudo masks in a fully supervised manner. We evaluate on PASCAL VOC 2012, a standard benchmark for weakly supervised methods, where Mask R-CNN trained on pseudo masks outperformed not only methods with the same level of supervision, image-level labels, but also methods that use counts and bounding boxes in their supervision.
\bibliography{egbib}
\end{document}